\documentclass[twoside]{article}

\newif\ifreview
\reviewfalse

\ifreview
\usepackage{template/aistats2023_author_response}
\else
\usepackage[accepted]{template/aistats2023}
\fi

\usepackage[T1]{fontenc}
\usepackage[english]{babel}

\usepackage{newtxtext}

\usepackage{mathtools} 
\usepackage{amsthm}
\allowdisplaybreaks    

\usepackage[nonewtxmathopt]{newtxmath} 
\usepackage[cal=cm,bb=ams,frak=pxtx]{mathalpha}

\usepackage{xcolor} 

\PassOptionsToPackage{hyphens}{url}
\usepackage[hyphens]{url}
\usepackage{hyperref}

\usepackage{graphicx}
\usepackage{caption}
\usepackage{tikz}
\usetikzlibrary{hobby,decorations.markings}
\usepackage{pgfplots}
\pgfplotsset{compat=1.17}

\usepackage{booktabs}
\usepackage{microtype}
\usepackage{csquotes}
\usepackage{enumitem}
\setitemize{itemsep=0pt,parsep=0pt,topsep=0pt}
\setenumerate{itemsep=0pt,parsep=0pt,topsep=0pt}
\usepackage[ruled,vlined]{algorithm2e}

\usepackage{xfrac}

\usepackage[round]{natbib}
\newcommand{\ind}[1]{\mathbf{1}_{#1}}

\DeclareMathOperator{\E}{\mathbb{E}}
\DeclareMathOperator{\Pbb}{\mathbb{P}}

\DeclareMathOperator*{\argmin}{arg\,min}

\DeclareMathOperator{\sign}{sign}
\DeclareMathOperator*{\supp}{supp}

\newcommand{\N}{\mathbb{N}}

\newcommand{\R}{\mathbb{R}}

\renewcommand{\brace}[1]{\left\{ #1 \right\}}
\newcommand{\bracket}[1]{\left[ #1 \right]}

\newcommand{\paren}[1]{\left( #1 \right)}
\newcommand{\midvert}{\,\middle\vert\,}

\newcommand{\abs}[1]{\left| #1 \right|}
\newcommand{\norm}[1]{\left\| #1 \right\|}
\newcommand{\scap}[2]{\left\langle #1, #2 \right\rangle}

\newcommand{\X}{\mathcal{X}}
\newcommand{\Y}{\mathcal{Y}}
\newcommand{\Z}{\mathcal{Z}}

\renewcommand{\epsilon}{\varepsilon}
\renewcommand{\phi}{\varphi}

\theoremstyle{plain}
\newtheorem{theorem}{Theorem}
\newtheorem{corollary}{Corollary}
\newtheorem{lemma}{Lemma}
\newtheorem{proposition}{Proposition}
\newtheorem{definition}{Definition}
\newtheorem{assumption}{Assumption}

\newtheorem{example}{Example}

\renewcommand\refname{References}

\begin{document}

\ifreview
\else
\twocolumn[
\aistatstitle{A Case of Exponential Convergence Rates for SVM}
\aistatsauthor{ Vivien Cabannes \And Stefano Vigogna }
\aistatsaddress{ Meta AI \And  University of Rome Tor Vergata } ]

\begin{abstract}
  
Optimizing the misclassification risk is in general NP-hard. 
Tractable solvers can be obtained by considering a surrogate regression problem. 
While convergence to the regression function is typically sublinear, the corresponding classification error can decay much faster. 
Fast and super fast rates (up to exponential) have been established for general smooth losses on problems where a hard margin is present between classes.
This leaves out models based on non-smooth losses such as support vector machines, and problems where there is no hard margin, begging several questions. Are such models incapable of fast convergence?
Are they therefore structurally inferior? Is the hard margin condition really necessary to obtain exponential convergence?
Developing a new strategy, we provide an answer to these questions.
In particular, we show not only that support vector machines can indeed converge exponentially fast, but also that they can do so even without hard margin.

\end{abstract}

\section{INTRODUCTION}
To solve a problem with computer calculations, classical computer science consists in handcrafting a set of rules.
In contrast, machine learning is based on the collection of a vast amount of solved instances of this problem, and on the automatic tuning of an algorithm that maps inputs defining the problem to the desired outputs.
Denote the inputs by $x \in \X$, the outputs by $y\in\Y$, and the input/output mappings by $f:\X\to\Y$.
To learn a mapping $f^*$, it is customary to introduce an explicit metric of error, and search for the function that minimizes it.
Define this metric through a loss $\ell:\Y\times\Y\to\R$ that quantifies how bad a prediction $f(x)$ is when $y$ is observed.
Assuming the existence of a distribution $\rho$ over $\X\times\Y$, that generates the instances of the problem meant to be solved, one aims to minimize the average loss value
\begin{equation}
	\label{svm:eq:obj}
	{\cal R}(f) = \E_{(X, Y)\sim\rho}\bracket{\ell(f(X), Y)}.
\end{equation}
In practice, this ``risk'' ${\cal R}$ can be evaluated approximately with samples ${\cal D}_n = (X_i, Y_i)_{i\leq n}$, collected by the machine learning scientist and assumed to have been drawn independently accordingly to $\rho$.

This work focuses on the binary classification problem where $\Y = \brace{-1, 1}$, and $\ell$ is the zero-one loss $\ell(y, z) = \ind{y\neq z}$.
In this setting, the risk ${\cal R}(f)$ captures the probability of mistakes of a classifier $f$, and its minimizer is characterized by
\begin{align}
	& f^* = \argmin_{f:\X\to\Y} {\cal R}(f) = \sign \eta, \\
	\text{where} \qquad & \eta(x) = \E\bracket{Y\midvert X=x}.
\end{align}
Ideally, leveraging the dataset ${\cal D}_n$, one would like to find a mapping $f_{{\cal D}_n}:\X\to\Y$ that is close to be optimal, in the sense that the excess of risk
\(
{\cal E}(f_{{\cal D}_n}) = {\cal R}(f_{{\cal D}_n}) - {\cal R}(f^*)
\)
is as small as it could be.
Since this quantity is actually random, inheriting from the randomness of the samples, statisticians focus on controlling its average.
While classification is often the first problem described in introductory machine learning classes, several recent works have shown that, when the model is well-specified, as the number of samples grows, it is possible to show that this average decays much faster than what usual statistical learning theory suggests.
This section provides a brief historical review of related literature before précising our contributions.

\subsection{Statistical Learning Theory}
The classical approach to minimize \eqref{svm:eq:obj} without the knowledge of $\rho$ but with the sole access to samples ${\cal D}_n \sim \rho^{\otimes n}$ is to restrict the search over functions in a class $\cal F$, and look for an empirical risk minimizer
\begin{align}
	\label{svm:eq:erm}
	& f^*_{{\cal D}_n} \in \argmin_{f\in{\cal F}} {\cal R}_{{\cal D}_n}, \\
	\text{where} \qquad & {\cal R}_{{\cal D}_n}(f) = \frac{1}{n} \sum_{i=1}^n \ell(f(X_i), Y_i). \nonumber
\end{align}
If we denote by $f^*_{\cal F}$ the minimizer of ${\cal R}$ in ${\cal F}$, using the fact that ${\cal R}_{{\cal D}_n}(f^*_{\cal F}) \geq {\cal R}_{{\cal D}_n}(f^*_{{\cal D}_n})$, the excess of risk can be bounded as
\begin{align}
	{\cal R}(f_{{\cal D}_n}^*) - {\cal R}(f^*)
	&\leq
	2\sup_{f\in{\cal F}} \abs{{\cal R}(f) - {\cal R}_{{\cal D}_n}(f)}
	\ \text{\scriptsize (estimation error)} \nonumber \\
	&+
	{\cal R}(f^*_{\cal F}) - {\cal R}(f^*) .
	\ \text{\scriptsize (approximation error)} 	\label{svm:eq:app_est}
\end{align}
This bound can be seen as highly suboptimal because it bounds the deviation of a random function with the worst deviation in the function class.
However, for any class~${\cal F}$, there exists an ``adversarial'' distribution $\rho$ for which convergence rates (of the excess of risk toward zero as a function of the number of samples $n\in\N$) derived through this bound can not be improved beside lowering some multiplicative constants \citep{Vapnik1995}.
On the one hand, the estimation error can be controlled with general tools to bound the supremum of a random process \citep[\emph{e.g.},][]{Dudley1967}, and will decrease as $O(n^{-1/2})$ with a multiplicative constant that depends on the size of the class ${\cal F}$.
On the other hand, the approximation error depends on assumptions of the problem, and the bigger the size of the class ${\cal F}$, the less restrictive it will be to assume that $f^*$ is not too different from $f_{\cal F}^*$.
Hence, there is a clear trade-off between controlling both errors, which should be balanced in order to optimize a bound on the full excess of risk.

\subsection{Surrogate Methods}
In practice, due to the combinatorial nature of discrete-valued functions, finding the empirical risk minimizer \eqref{svm:eq:erm} is often an intractable problem \citep[\emph{e.g.},][]{Hoffgen1992,Arora1997}.
Therefore, people have approached the original problem with other perspectives.
A straightforward approach is given by \emph{plug-in classifiers}, \emph{i.e.}, classifiers of the form $\sign \hat\eta$, for $\hat\eta$ some estimator of $\eta$.
For example, such an estimator can be constructed as
\(
\hat\eta(x) = \sum_{i=1}^n \alpha_i(x) Y_i,
\)
for $\alpha_i(x)$ some weights that specify how much the observation $Y_i$ made at the point $X_i$ should diffuse to the point $x$ \citep[see][for an example]{Friedman1994}.
Another popular approach to solve classification problems is provided by \emph{support vector machines} (SVM), which were introduced from geometric considerations to maximize the margin between the classes $\brace{x\in\X\midvert f^*(x) = y}$ for $y \in\brace{-1, 1}$ \citep{Cortes1995}.

These two approaches can be conjointly understood as introducing a surrogate loss $L:\R\times\Y\to\R$ and looking for a continuous-valued function $g:\X\to\R$ that solves the surrogate problem defined by $L$ before retrieving a solution $f:\X\to\Y$ of the original problem as
\begin{align}
	\label{svm:eq:decoding}
	& f = \sign g, \qquad
	g^*\in \argmin_{g:\X\to\R} {\cal R}_S(g), \\
	\text{where}\qquad
	& {\cal R}_S(g) = \E_{(X,Y)\sim\rho}\bracket{L(g(X), Y)}, \nonumber
\end{align}
where the notation $S$ stands for ``surrogate''.
To an estimate $g:\X\to\R$ of $g^*$ we associate an estimate $f:\X\to\Y$ of $f^*$ through the decoding step $f=\sign g$.
In particular, using the variational characterization of the mean, $\eta$ can be estimated through $L(z, y) = \abs{z - y}^2$.
Meanwhile, SVM are related to the \emph{hinge loss}, see Appendix \ref{svm:app:hinge_loss},
\begin{equation}
	\label{svm:eq:Hinge}
	L(z, y) = (1-zy)_+ = \max\paren{0, 1 - zy},
\end{equation}
Surrogate methods benefit from their relative easiness to optimize and the quality of their practical results.
Arguably, they define the current state of the art in classification, softmax regression being particularly popular to train neural networks on classification tasks.

Surrogate methods were studied in depth by \cite{Zhang2004,Bartlett2006}, who proposed a generic framework to relate the excess of the original risk to the excess of surrogate risk through calibration inequalities of the type
\begin{equation}
	\label{svm:eq:calibration}
	{\cal R}(f) - {\cal R}(f^*) \leq \psi\paren{{\cal R}_S(g) - {\cal R}_S(g^*)},
\end{equation}
where $f = \sign g$
and $\psi$ is a concave function, uniquely defined from $L$ and verifying $\psi(0) = 0$.
The use of a concave function is motivated by Jensen inequality, allowing to integrate an inequality derived pointwise (conditionally on an input $x$).

When $\psi(x) \propto x$ and $\E_{{\cal D}_n}{\cal R}_S(g_{{\cal D}_n}) - {\cal R}_S(g^*)$ is controlled as $O(n^{-1/2})$ for $g_{{\cal D}_n}$ the minimizer of the empirical surrogate risk, one can control $\E_{{\cal D}_n}{\cal R}(\sign g_{{\cal D}_n}) - {\cal R}(f^*)$ as $O(n^{-1/2})$.
Because of minimax optimality of VC theory, those rates in $n^{-1/2}$ are wildly regarded as optimal. 
Yet, a closer look reveals that examples to show minimax optimality are based on “degenerated” data distributions. 
Those worst cases are unlikely to appear in practice. 
Indeed, some works have shown that under relatively mild low-noise conditions, much faster rates can be derived. 

\subsection{Exponential Convergence Rates}
On the one hand, calibration inequalities \eqref{svm:eq:calibration} are appealing, as they allow casting directly rates derived on the surrogate problem to rates on the original problem.
On the other hand, because~$\psi$ has to be concave, rates in $O(n^{-r})$ on the surrogate problem can not be cast as better rates on the original problem, corresponding to the optimal inequalities where $\psi(x) = cx$ for some $c > 0$,
Yet, one can find cases where the sign of $\eta$ can be estimated much faster than $\eta$ itself, even when this sign is estimated with surrogate methods.

Most works that prove faster convergence rates are based on a specific derivation, which consists in relating the classification excess of risk with some power of the supremum norm.
More specifically, \cite{Mammen1999} \citep[see also][]{Massart2006} introduced the following condition.

\begin{assumption}[Hard low-noise condition]
	\label{svm:ass:margin}
	The binary classification problem defined through the distribution $\rho$ is said to verify the hard low-noise (or hard Tsybakov margin\footnote{Note that the wording margin here refers to a form of separation in the output space, and not to the usual margin of SVM that describes separation between classes in the input space.}) condition if the conditional mean $\eta$ is bounded away from zero, i.e.,
	\begin{equation}
		\label{svm:eq:margin}
		\exists\,\eta_0 > 0; \qquad \abs{\eta(X)} > \eta_0\qquad\textit{a.s.},
	\end{equation}
	where the notation a.s. stands for almost surely.
	Equivalently, $\abs{\eta}^{-1} \in L^\infty(\rho_\X)$.
\end{assumption}

Under Assumption \ref{svm:ass:margin}, we know that $\abs{g(x) - \eta(x)} < \abs{\eta(x)}$ implies $\sign g(x) = \sign \eta(x) = f^*(x)$, thus ${\cal R}(\sign g) - {\cal R}(f^*) \leq \Pbb_X(\sign g(X) \neq f^*(X)) \leq 1_{\norm{g - \eta}_{\infty} \geq \eta_0}$.
Hence, we get for any estimate $g_{{\cal D}_n}:\X\to\R$ computed from the dataset ${\cal D}_n$,
\[
	\E_{{\cal D}_n}[{\cal R}(\sign g_{{\cal D}_n})] - {\cal R}(f^*)
	\leq \Pbb_{{\cal D}_n}\paren{\norm{g_{{\cal D}_n} - \eta}_{L^\infty} > \eta_0}.
\]
As a consequence, an exponential concentration inequality on the $L^\infty$ distance between $g_{{\cal D}_n}$ and $\eta$ directly translates to exponential convergence rates on the average excess of risk.
In particular, estimation methods for~$\eta$ based on H\"older classes of functions, such as local polynomials, are known to be well-behaved with respect to the $L^\infty$ norm \citep[see, \emph{e.g.}, the construction of covering number by][]{Kolmogorov1959}.
This was leveraged by \cite{Audibert2007} in a seminal paper that shows how better rates can be achieved on the classification problem under Assumption \ref{svm:ass:margin} and a variety of weaker conditions (described later in Assumption \ref{svm:ass:weak_margin}).

Surprisingly, such an approach has remained somehow less popular than approaches based on calibration inequalities, and we are missing a framework to fully apprehend fast rates phenomena.
Some results for logistic regression were achieved by \cite{Koltchinskii2005}.
Recently, \cite{Cabannes2021b} showed that this result generalizes to any discrete output learning problem, and that approaches that naturally lead to concentration in $L^2$ could be turned into fast rates based on interpolation inequalities that relate the $L^2$ norm with the $L^\infty$ one (notably reusing the work of \cite{Fischer2020} on interpolation spaces).
Exploiting the work of \cite{MarteauFerey2019}, this can be generalized to any self-concordant loss (using self-concordance to reduce the problem to a least-squares problem); and, through the work of \cite{Lin2020}, to any spectral filtering technique (beyond Tikhonov regularization), such as stochastic gradient descent, which was actually shown earlier for binary classification by \cite{PillaudVivien2018b} and \cite{Nitanda2019}.
In the same stream of research, \cite{Vigogna2022} proposed a general framework to study exponential rates for smooth losses in multiclass classification beyond least-squares.
However, we believe that there is a bigger picture to be uncovered.

\subsection{Contribution}
The proofs of exponential convergence in the works quoted above are all based on the basic mechanism outlined in \cite{Audibert2007}, which, in substance, consists in relating the excess of classification risk with concentration on the supremum norm as, with notations of Assumption \ref{svm:ass:weak_margin},
\[{\cal R}(\sign g) - {\cal R}(\sign g^*) \leq c\norm{g - g^*}_{\infty}^{p+1},\]
Unfortunately, such a mechanism relies on $L^\infty$ concentration, and does not easily extend to the hinge loss.
Does this mean that support vector machines do not exhibit superfast rates, and thus they are inferior to other surrogate methods?
The practice seems to answer negatively.
In this paper, we give a firm theoretical answer to this question.
In particular, we show not only that support vector machines do achieve exponential rates, but also that they can do so even without assuming the hard low-noise condition.
Our main contribution is to introduce a general framework to prove exponential convergence rates, and show how this framework can be applied to the hinge loss while only considering classical assumptions.
Stated otherwise, our goal is to show fast rates for a non-smooth loss, which we will do for the hinge loss formalization of kernelized SVM.

\paragraph{Outline.}
Our general strategy is illustrated on Figure \ref{svm:fig:discrete_analysis} and consists in first finding a relation
\[
	{\cal R}_S(g_{\theta}) - {\cal R}_S(g_{\theta^*}) \geq \norm{\theta - \theta^*},
\]
for some natural parameter $\theta$ in a Banach space $\Theta$ parametrizing a class of functions $g_\theta\in{\cal F}$, and then show that $\sign g_\theta = \sign g_{\theta^*}$ when $\norm{\theta - \theta^*}$ is small enough, that is,
\[
	\exists\, \epsilon > 0; \qquad
	\norm{\theta - \theta^*} \leq \epsilon \quad\Rightarrow\quad
	\sign g_\theta = \sign g_{\theta^*}.
\]
Assuming that $\sign g_{\theta^*} = f^*$, we deduce that
\begin{align*}
&\E_{{\cal D}_n}[{\cal R}(\sign g_{\theta_n})] - {\cal R}(f^*) 
\\= &\E_{{\cal D}_n}[{\cal R}(\sign g_{\theta_n}) - {\cal R}(\sign g_{\theta^*})]
\\\leq& \E_{{\cal D}_n}[1_{\sign g_{\theta_n} \neq (\sign g_{\theta^*})}]
\\\leq& \E_{{\cal D}_n}[1_{\norm{\theta_n - \theta^*}\geq \epsilon}]
\\\leq& \Pbb_{{\cal D}_n}(\norm{\theta_n - \theta^*}\geq \epsilon)
\\\leq& \Pbb_{{\cal D}_n}({\cal R}_S(g_{\theta_n}) - {\cal R}_S(g_{\theta^*})\geq \epsilon).
\end{align*}
where $g_{\theta_n}$ is an estimate of $g_\theta$ based on the samples ${\cal D}_n$.
Finally, we conclude with an exponential concentration inequality that controls the deviation of the excess of risk based on classical statistical learning theory.

\begin{figure*}
	\centering

\begin{tikzpicture}
  
  \filldraw[yellow] (1.9, 0) circle (1);

  \draw[dashed] (0, 0) ellipse (.5 and .5);
  \draw[dashed] (.2, 0) ellipse (.9 and .75);
  \draw[dashed] (.4, 0) ellipse (1.3 and 1);
  \draw[dashed] (.6, 0) ellipse (1.7 and 1.25);
  \draw[densely dotted,red,thick] (.8, 0) ellipse (2.1 and 1.5);
  \draw[dashed] (1, 0) ellipse (2.5 and 1.75);

  \begin{scope}[decoration={markings, 
      mark= at position .15 with {\arrow{latex}},
      mark= at position .35 with {\arrow{latex}},
      mark= at position .55 with {\arrow{latex}},
      mark= at position .75 with {\arrow{latex}},
      mark= at position .95 with {\arrow{latex}}}
    ]
    \draw [postaction={decorate},gray,thick] (4,0) -- (0,0);
  \end{scope}

  \filldraw (0, 0) circle (2pt) node[anchor=north west] {$\theta^*$};
  \filldraw (1.9, 0) circle (2pt) node[anchor=north west] {$\theta_\lambda$};

  \draw[dashed] (4, 2) -- (5, 2) node[anchor=west] {level lines of ${\cal R}(\sign g_{\theta_n})$};
  \begin{scope}[decoration={markings, 
      mark= at position .5with {\arrow{latex}}}
    ]
    \draw[gray,thick,postaction={decorate}] (5, 1.5) node[anchor=west,black] {path $\brace{\theta_\lambda; \lambda>0}$} -- (4, 1.5);
  \end{scope}
  \filldraw[yellow] (4.5, 1) circle (.25);
  \draw (5, 1) node[anchor=west,black]{region $\brace{\theta; \|\theta - \theta_\lambda \| < 1}$};
 
  \draw[densely dotted,red,thick] (4, .5) -- (5, .5) node[anchor=west,black] {certified value of ${\cal R}(\sign g_{\theta_n})$ when $\|\theta_n - \theta_\lambda\| < 1$};

\end{tikzpicture}
	\caption{Our convergence analysis consists in relating natural concentration given by surrogate methods to the original excess of risk without passing by the surrogate excess of risk.
	We denote by $\theta_\lambda$ the regularized surrogate risk minimizer $\argmin {\cal R}_S(g_\theta) + \lambda\norm{\theta}^2$.
	As the drawing shows, concentration in parameter space $\Theta$ can be cast as deviation on the original excess of risk.
	Yet, such a casting relation depends on the geometry of this picture, which itself depends on what surrogate is used, what is the function to learn, how a regularized estimator approached it, and how our empirical estimate concentrates around the regularized estimator.
	Note that this figure illustrates an abstract mechanism that generalizes the simpler mechanism we use to derive exponential convergence rates.
	It contrasts with usual statistical learning theory that combines approximation and estimation error in an additive fashion.
	}
	\label{svm:fig:discrete_analysis}
\end{figure*}
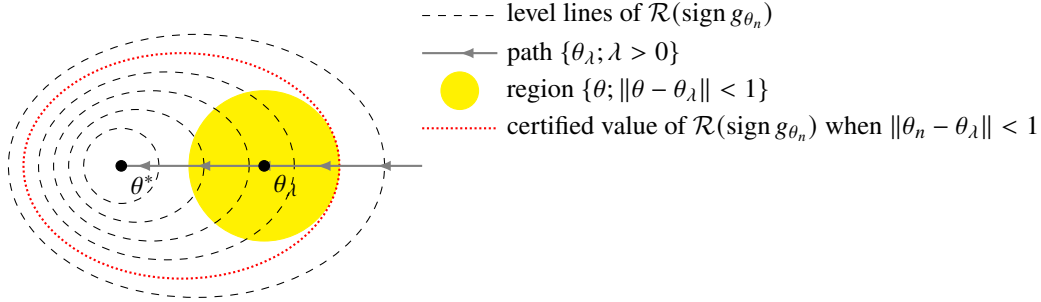

\section{EXPONENTIAL CONVERGENCE OF SVM}
This section is devoted to the proof of exponential convergence rates for the hinge loss.
We shall fix the notation ${\cal R}_S$ as the surrogate risk associated with \eqref{svm:eq:Hinge}.
All the proofs are collected in Appendix \ref{svm:app:proofs}.

\subsection{Refined Calibration for the Hinge Loss}

We start by introducing the classical weak low-noise condition \citep{Mammen1999}.

\begin{assumption}[Weak low-noise condition]
	\label{svm:ass:weak_margin}
	The binary classification problem defined through the distribution $\rho$ is said to verify the $p$-low-noise condition, with $p \in (0, \infty)$, if there exists a constant $c > 0$ such that
	\begin{equation}
		\Pbb_{\rho_\X}(0 < \abs{\eta(X)} < t) \leq c t^p,
	\end{equation}
	where the notation $\rho_\X$ denotes the marginal of $\rho$ over $\X$.
\end{assumption}

Assumption \ref{svm:ass:weak_margin} is equivalent to asking for the inverse of the conditional mean~$\abs{\eta}^{-1}$ (with the convention $0^{-1} = 0$) to belong to the Lorentz space $L^{p,\infty}(\rho_\X)$ (also known as weak-$L^p$ space), which is the Banach space endowed with the norm (quasi-norm and quasi-Banach if $p < 1$)
\begin{equation}
	\norm{f}_{p, \infty} = \sup_{t > 0} t\Pbb_{\rho_\X}(f(X) > t)^{\frac{1}{p}},
\end{equation}
where the $\rho_\X$ denotes the marginal of $\rho$ with respect to $\X$.
This definition can be extended to the case $p=\infty$ by setting $ L^{p,\infty}(\rho_\X) = L^\infty(\rho_\X) $, which characterizes the hard low-noise condition in Assumption \ref{svm:ass:margin}.
We will also use~$\norm{\cdot}_p$, for $p\in[1,\infty]$, to denote the $L^p$-norm on $\X$ endowed with~$\rho_\X$.

We now relate the excess of risk on the hinge loss to the deviation in these spaces.

\begin{lemma}[Weak-$L^q$ concentration due to the hinge loss]
	\label{svm:lem:l1}
	For any functions $g_1, g_2:\X\to[-1,1]$,
	\begin{equation} \label{svm:lem1a}
		{\cal R}_S(g_2) - {\cal R}_S(g_1) = \E_{\rho_\X}[ -\eta(X) (g_2(X) - g_1(X))].
	\end{equation}
	In particular, under Assumption \ref{svm:ass:margin}, for any $g:\X\to\R$,
	\begin{equation} \label{svm:lem1b}
		{\cal R}_S(g) - {\cal R}_S(g^*) \geq \norm{\abs{\eta}^{-1}}_{\infty}^{-1} \norm{\pi(g) - g^*}_{1}.
	\end{equation}
	where $g^* = \sign \eta$ is a minimizer of ${\cal R}_S$ and $\pi$ is the projection of $\R$ on $[-1, 1]$, defined as mapping $t\in\R$ to $\pi(t) = \sign(t) \min\{|t|,1\}$.
	Similarly, under Assumption \ref{svm:ass:weak_margin}, with $q = \sfrac{p}{p+1}$,
	\begin{equation} \label{svm:lem1c}
		{\cal R}_S(g) - {\cal R}_S(g^*)
		\geq 2^{-1} \norm{\abs{\eta}^{-1}}_{p,\infty}^{-1} \norm{\pi(g) - g^*}_{q,\infty}.
	\end{equation}
\end{lemma}

Lemma \ref{svm:lem:l1} shows that we can set the minimizer $g^* = f^* \in \brace{-1, 1}^\X$.
This is a useful fact as it implies that the excess of the original risk is zero as soon as $\norm{g - g^*}_\infty < 1$.
In essence, the only piece missing in order to prove fast convergence rates is an interpolation inequality between $L^{q,\infty}$ and $L^\infty$.
In the following, we will leverage Lemma \ref{svm:lem:l1} more subtly by considering a class of functions ${\cal G}$ and assumptions on the distribution $\rho_\X$ such that, if an estimate $g\in{\cal G}$ has not the same sign almost everywhere as the estimand $g^*$, then $\norm{g-g^*}_{q, \infty}$ is bounded away from zero.
By contraposition, if $g\in{\cal G}$ presents a small excess of surrogate risk, then $\sign g = \sign g^*$.
When $\X$ is a metric space, one way to proceed is to assume that $g$ is Lipschitz-continuous, together with some minimal mass assumptions.
Let us begin with the minimal mass assumption.
We first need the following definition.

\begin{definition}[Well-behaved sets]
	A set $U\subset\X$ is said to be well-behaved with respect to $\rho$ if there exist constants $c, r > 0$ and an exponent $d > 0$ such that, for any $x\in U$,
	\begin{equation}
		\label{svm:eq:mass}
		\forall\,\epsilon \in [0, r]; \qquad \rho_\X(U\cap{\cal B}(x, \epsilon)) \geq c\epsilon^d,
	\end{equation}
	and ${\cal B}(x, \epsilon)$ the ball in $\X$ of center $x$ and radius $\epsilon$.
\end{definition}

The following examples show that the coefficient~$d$ that appears in \eqref{svm:eq:mass} results from the dimension of the ambient space, the regularity of singularities of the border of the set, and the decay of the density when approaching the frontier of the set.

\begin{example}
	The set $[0, 1]^p$ is well-behaved with coefficients $r=1$, $d=p$ and $c=2^{-d}\operatorname{vol}(\mathbb{S}^{d-1})$ with respect to the Lebesgue measure in $\R^d$.
\end{example}
\begin{example}
	The set $\brace{(x, y) \in \R^2\midvert x\in[0,1], y\in[0, x^{n-1}]}$ is well-behaved with coefficient $r=1$, $d=n$ and $c=n^{-1}$ with respect to the Lebesgue measure.
	Reciprocally, the set $[0,1]$ is well-behaved with coefficient $r=1$, $d=n$ and $c=n^{-1}$ with respect to the measure whose density equals $p(x) = x^{n-1}$.
\end{example}

\begin{assumption}[Minimal mass assumption]
	\label{svm:ass:mass}
	The classification problem is said to verify the $d$-minimal mass assumption if the decision regions $\X_y = \brace{x\in\supp \rho_\X\midvert y\eta(x) > 0}$ for $y \in \brace{-1, 1}$ are both well-behaved with exponent $d$.
\end{assumption}

Assumption \ref{svm:ass:mass} is a weakening of an assumption that is commonly found in the statistical learning literature.
More precisely, it is often assumed that $\rho$ is absolutely continuous according to the Lebesgue measure $\lambda$ on $\X$ (assumed to be a Euclidean space), that its density is bounded away from zero on its support, and that its support has smooth boundary, so that $\lambda(\supp\rho_\X\cap {\cal B}(x, \epsilon)) > c'\lambda({\cal B}(x, \epsilon))$ \citep[see the strong density assumption in][]{Audibert2007}.

The minimal mass requirement allows relating misclassification events to $L^{q,\infty}$ deviation.

\begin{lemma} \label{svm:lem2}
	Under Assumption \ref{svm:ass:mass}, there exists a constant $c_0$ such that if $g$ is $G$-Lipschitz-continuous for $G > r^{-1}$, for any $q \in (0, 1]$
	\begin{equation}
	\begin{aligned}
		& \exists\, x \in \supp\rho_\X,\,\,\, \abs{g(x) - g^*(x)} \geq 1 \\
		\Rightarrow \quad & \norm{g - g^*}_{q, \infty} \geq c_0 G^{-\frac{d}{q}}.
		\end{aligned}
	\end{equation}
\end{lemma}

Putting together Lemmas~\ref{svm:lem:l1} and~\ref{svm:lem2}, we obtain the following refined calibration.

\begin{proposition}
	\label{svm:prop:mid}
	Under Assumptions \ref{svm:ass:weak_margin} and \ref{svm:ass:mass}, if $g$ is $G$-Lipschitz-continuous with $G>r^{-1}$, we have
	\begin{equation}
	\begin{aligned}
		\label{svm:eq:mid}
		& {\cal R}_S(g) - {\cal R}_S(g^*) \leq 2^{-1}\norm{\abs{\eta}^{-1}}_{p,\infty}^{-1} c_0 G^{-\frac{d(p+1)}{p}} \\
		\Rightarrow \quad & {\cal R}(\sign g) = {\cal R}(f^*).
	\end{aligned}
	\end{equation}
\end{proposition}

\subsection{Trade-off between Estimation and Approximation Errors}
We are now left with the research of $g_{{\cal D}_n}$ inside a class of Lipschitz-continuous functions such that ${\cal R}_S(g_{{\cal D}_n}) - {\cal R}_S(g^*)$ is sub-Gaussian (its randomness being inherited from the dataset ${\cal D}_n$ from which $g_{{\cal D}_n}$ is built).
To do so, let us consider a linear ({\em a.k.a.} kernelized) class of functions
\begin{equation}
	\label{svm:eq:f_k}
	{\cal G}_{M, \sigma} = \brace{x\mapsto\scap{\theta}{\phi(\sigma^{-1} x)}\midvert \theta \in {\cal H}, \norm{\theta}_{\cal H} \leq M} ,
\end{equation}
where ${\cal H}$ is a separable Hilbert space, $\phi:\X\to{\cal H}$ is a $G_\phi$-Lipschitz-continuous mapping, and $\sigma > 0$ is a scaling (or bandwidth) parameter.
Such a class of functions can be entirely described from the kernel $k(x, x') = \scap{\phi(x)}{\phi(x')}$ \citep[see][for a primer on kernel methods]{Scholkopf2001}.
An example for ${\cal G}$ is given by the Gaussian kernel, a.k.a. radial basis function, $k(x, x') = \exp(-d(x, x')^2)$.
Using Cauchy-Schwarz, it is easy to show that any function in ${\cal G}_{M,\sigma}$ is $MG_\phi \sigma^{-1}$-Lipschitz-continuous.

In order to find a function $g_{{\cal D}_n}$ that is likely to minimize ${\cal R}_S$ without accessing the distribution $\rho$, but only i.i.d. samples ${\cal D}_n = (X_i, Y_i)_{i\leq n} \sim \rho^{\otimes n}$, it is classical to consider the empirical risk minimizer
\begin{equation}
	\label{svm:eq:sur_erm}
	g_{{\cal D}_n} \in \argmin_{g\in{\cal G}_{M, \sigma}} \frac{1}{n} \sum_{i=1}^n L(g(X_i), Y_i).
\end{equation}
This problem is convex with respect to $\theta$ parametrizing $g \in {\cal G}_{M,\sigma}$, and is easily optimized with duality.
We refer the curious reader to the extensive literature on kernelized SVM \citep[see][for books on the matter]{Cristianini2000,Scholkopf2001,Steinwart2008}.

In order to show that ${\cal R}_S(g_{{\cal D}_n})$ is close to ${\cal R}_S(g^*)$, one can apply classical results from statistical learning theory, and in particular \eqref{svm:eq:app_est}.
The estimation error can be bounded using the extensive literature on Rademacher complexity for linear classes of functions on Lipschitz-continuous losses \citep{Bartlett2002}.
To bound the approximation error, one needs to make additional assumptions on the problem.
We refer to \cite{Steinwart2007,Blaschzyk2018} for advanced considerations on the matter.
In view of our calibration result \eqref{svm:eq:mid}, the following additional assumption suffices to prove exponential convergence of SVM.

\begin{assumption}[$(c_0, p,d)$-Source condition]
	\label{svm:ass:source}
	There exist $M, \sigma$ and a function $g\in{\cal G}_{M, \sigma}$ such that ${\cal R}_S(g) - {\cal R}_S(g^*) \leq 4^{-1}\|\abs{\eta}^{-1}\|_{p, \infty}^{-1} c_0 M^{-r} G_\phi^{-r} \sigma^r$ with $r = d(p+1) / p$.
\end{assumption}

It should be noted that, because of Proposition \ref{svm:prop:mid}, the function $g$ in Assumption \ref{svm:ass:source} is a perfect classifier. This implies that the decision frontier $\overline{\X_{-1}} \cap \overline{\X_1}$ (the bar notation corresponding to space closure) inherits from the regularity of $g$, since it is included in the set $\brace{x\in\X\midvert g(x) = 0}$.
In particular, if ${\cal G}_{M,\sigma}$ is included in ${\cal C}^m$, this frontier would be in ${\cal C}^m$.
Hence, for Assumptions \ref{svm:ass:source} to hold, the boundary frontier should match the regularity implicitly defined by ${\cal G}_{M, \sigma}$.

We are finally ready to state our main result, establishing exponential convergence rates for SVM.

\begin{theorem}[Exponential convergence rates for SVM]
	\label{svm:thm}
	Under Assumptions \ref{svm:ass:weak_margin}, \ref{svm:ass:mass} and \ref{svm:ass:source}, there exists a constant $c > 0$ such that the empirical minimizer $g_{{\cal D}_n}$ defined by \eqref{svm:eq:sur_erm} verifies
	\begin{equation}
		\E_{{\cal D}_n}{\cal R}(\sign g_{{\cal D}_n}) - {\cal R}(f^*)
		\leq 2\exp(-cn).
	\end{equation}
\end{theorem}

\begin{figure*}[t]
	\centering
	\includegraphics{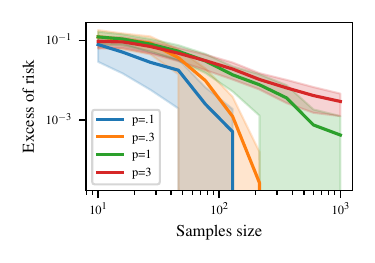}
	\includegraphics{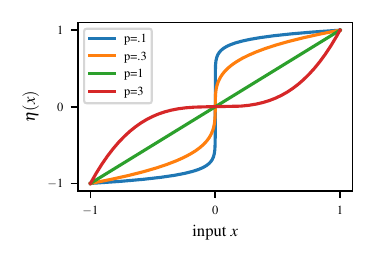}
	\caption{
		SVM generalization error as a function of the number of samples (left) for a problem where $X$ is uniform on $[-1, .-1] \cup [.1, 1]$ and $\eta(x) = \sign(x) \abs{x}^p$ (right). We observe exponential convergence rates on the blue and orange curve. The green and red curves showcase regimes where exponential convergence rates have not been reached yet, one would need more samples to observe them.
	}
	\label{svm:fig:fig1}
\end{figure*}

While this result is achieved for constrained SVM, the same result can be achieved with regularized SVM, which is used in practice.

\begin{corollary}
    \label{thm:cor}
    Under Assumptions \ref{svm:ass:weak_margin}, \ref{svm:ass:mass} and \ref{svm:ass:source}, there exist $ \lambda \geq 0 $ and a constant $c > 0$ such that $g_{{\cal D}_n, \lambda}$ verifies
	\begin{equation*}
		\E_{{\cal D}_n}{\cal R}(\sign g_{{\cal D}_n, \lambda}) - {\cal R}(f^*)
		\leq 2\exp(-cn).
	\end{equation*}
	where
    \[
        g_{{\cal D}_n, \lambda} = \argmin_{g_\theta; \theta\in{\cal H}} \sum_{i=1}^n L(g_\theta(X_i), Y_i) + \lambda\norm{\theta}_{\cal H}^2 .
    \]
\end{corollary}

\subsection{Relaxing Assumptions}

Exponential convergence rates rely on strong assumptions in order to set the approximation error to zero.
In particular, it is customary to assume that the surrogate function to learn lies in the model we have chosen, that is, in our notation, $g^* \in {\cal G}_{M, \sigma}$.
In our case this would be a strong assumption, since $g^*$ is piecewise constant while ${\cal G}_{M, \sigma}$ is a smooth space of functions.
It turns out that the assumption $g^*\in{\cal G}_{M, \sigma}$ is not necessary, and what we actually need is the ability to reach a sufficiently small risk within the class ${\cal G}_{M, \sigma}$.
How small is enough is quantified by the statement of Proposition \ref{svm:prop:mid} and Assumption \ref{svm:ass:source}.
In particular, this assumption is verified when the function class ${\cal G}_{M,\sigma}$ is rich enough and classes are separated by a margin in the input space as specified by the following assumption.

\begin{assumption}[Cluster Assumption]
    \label{svm:ass:cluster}
    The classes $\X_1 = (f^*)^{-1}(1) := \brace{x\in\supp\rho_\X\midvert \eta(x) >0}$ and $\X_{-1} = (f^*)^{-1}(-1)$ are separated by a margin, in the sense that the distance between any two points in each set is bounded away from zero. Formally,
    \begin{equation}
        \exists\,\delta_0 > 0;\quad\forall\, (x, x') \in \X_1\times \X_{-1},\qquad d(x, x') \geq \delta_0.
    \end{equation}
\end{assumption}

\begin{proposition}[Source condition example]
    \label{svm:prop:cluster}
    Assumption \ref{svm:ass:cluster} implies Assumption \ref{svm:ass:source} for any $(c_0, p, d)$, as long as $\X$ is a Euclidean space and $k$ is taken as the exponential kernel $k(x, y) = \exp(-\sigma^{-1}\norm{x-y})$ for any $\sigma > 0$.
\end{proposition}

In terms of practical applications, the cluster assumption says that no one can continuously modify an input to go from a region of the space linked with one class to a region linked with another class without going through inputs that will never exist.
This is typically true for well-curated image datasets such as CIFAR10: one can not continuously transform an image of a truck into an image of a horse without going through images that will never appear in the CIFAR10 dataset \citep{Krizhevsky2009}.
As stated in the seminal work of \cite{Seeger2001}, ``the `cluster assumption' is a very general and weak assumption, therefore applicable as prior assumption to many unsupervised tasks''.
It has been popular in unsupervised, weakly-supervised and semi-supervised learning \citep[see][for exponential convergence rates in those settings]{Rigollet2007,Cabannes2021}.

To deepen the study of the approximation error, one could leverage the following geometrical characterization of the risk of misclassification.
For $f:\X\to\brace{-1, 1}$, we have
\begin{equation}
\begin{aligned}
	\label{svm:eq:pre_source}
	& {\cal R}(f) - {\cal R}(f^*) = \E[\abs{\eta(X)} \ind{f(x) \neq f^*(X)}] \\
	\leq
	& \Pbb(f(X)\neq f^*(X)) = \rho_\X\paren{f^{-1}(\{1\})\,\triangle\, \X_1},
	\end{aligned}
\end{equation}
where $\triangle$ denotes the symmetric difference of sets, {\em i.e.} $A \triangle B = (A \cup B) \setminus (A \cap B)$.
In particular, under Assumption \ref{svm:ass:cluster} the minimizer $g_{{\cal G}_{M,\sigma}}$ of the surrogate risk in ${\cal G}_{M,\sigma}$ verifies
\begin{equation}
	\label{svm:eq:source}
	\rho_\X\paren{(\sign g_{{\cal G}_{M,\sigma}})^{-1}(\{1\})\,\triangle\, \X_1} \leq \psi(M, \sigma),
\end{equation}
for $\psi$ a function that vanishes for sufficiently large $M$ and small $\sigma$.

On the one hand, one could control the approximation error by assuming or deriving inequalities akin to \eqref{svm:eq:pre_source} and \eqref{svm:eq:source}, with different profiles of $\psi$.
We conjecture that this can be done by assuming low-noise conditions that are well adapted to the geometric nature of SVM, such as the one proposed by \cite{Steinwart2007} \citep[see also][]{Gentile1999,Cristianini2000}.
On the other hand, the estimation error can be controlled by extending the ideas presented in this paper to study the worst value of the estimation error ${\cal R}(\sign g_{{\cal D}_n}) - {\cal R}(\sign g_{{\cal G}_{M,\sigma}})$ under the knowledge of ${\cal R}_S(g_{{\cal D}_n}) - {\cal R}_S(g_{{\cal G}_{M,\sigma}})$.
Fitting $M$ and $\sigma$ to trade estimation and approximation errors, such derivations would open the way to fast polynomial rates under less restrictive assumptions.

\begin{figure*}[t]
	\centering
	\includegraphics{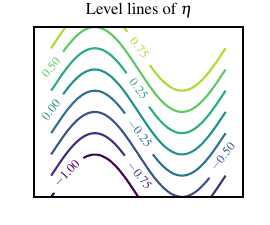}
	\includegraphics{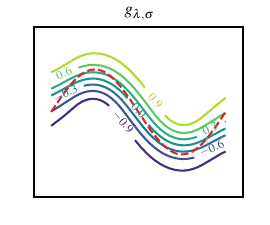}
	\includegraphics{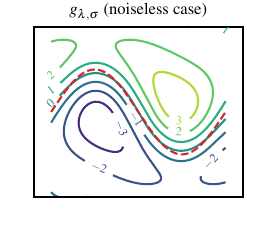}
	\includegraphics{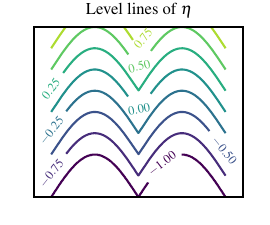}
	\includegraphics{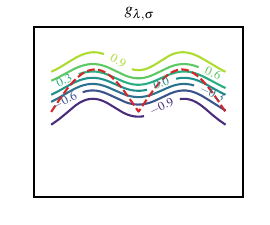}
	\includegraphics{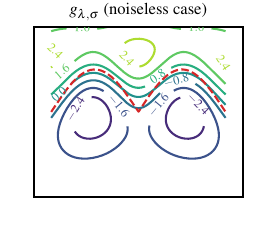}
	\caption{
	Study of the level lines of $g_{\lambda, \sigma}$ when $\eta^{-1}(0) \in {\cal C}^\infty$ (top) and $\eta^{-1}(0)\in{\cal C}^0\setminus {\cal C}^1$ (bottom).
	The function $g^*$ takes values $-1$ below the optimal decision frontier plotted in red and $+1$ above, independently of the noise.
	We observe that the bias error ${\cal R}(\sign g_{\lambda,\sigma}) - {\cal R}(f^*)$, which is bounded by the volume between the level lines $\brace{x\in\X\midvert g_{\lambda,\sigma}(x) = 0}$ and $\brace{x\in\X\midvert \eta(x) = 0}$ (plotted in red), depends on both the regularity of the latter, and on the noise level.
	Here, $\sigma$ is taken to be of the order of 15\% of the diameter of the domain, which explains the regularity of the observed level lines.
	The noiseless cases on the right correspond to the situations where $\E[Y\vert X] = \sign \eta(X)$ for $\eta$ plotted on the left.
	}
	\label{svm:fig:fig2}
\end{figure*}

\subsection{Prior Assumptions}
\label{svm:app:ass}
In this section, we discuss the mildness of our assumptions in comparison to existing proofs of exponential convergence rates.
Those proofs are based on the combination of the hard low-noise condition together with the existence of a regular function close to $\eta$, which implies Assumption \ref{svm:ass:cluster}.
In contrast, {\em the cluster assumption is both stronger than our assumptions and weaker than the existing assumptions to prove exponential convergence rates}. 

To our knowledge, \cite{Audibert2007} first proves exponential convergence rates. They did so by assuming the regression function $\eta$ to be H\"older-continuous, together with the hard Tsybakov condition (see Proposition 3.7 and the class ${\cal P}$).
Under the hard Tsybakov condition, $\X_1 = \eta^{-1}([\eta_0, +\infty))$, hence, for $x \in \X_1$ and $x' \in \X_{-1}$, if $\eta$ is $(L, \alpha)$-H\"older
\[
    2\eta_0 \leq \norm{\eta(x) - \eta(x')} \leq L\norm{x - x'}^\alpha.
\]
It follows that
\[
    d(\X_1, \X_{-1}) \geq \paren{\frac{2\eta_0}{L}}^{1/\alpha}.
\]
In other terms, those two assumptions implies the cluster assumption \ref{svm:ass:cluster}.
Similarly, \cite{Rigollet2007} proved exponential convergence rates under the cluster assumption for semi-supervised learning.

More recently, a renewed interest was triggered by results for SGD achieved by \cite{PillaudVivien2018b}.
Once again, the authors assumed both the hard Tsybakov margin condition (A1) and the existence of a perfect classifier that is in the linear space of functions considered (implied by A4). 
They discuss the fact that their assumptions are met under the cluster assumption (A5) plus some regularity of $\eta$ (see Proposition 3).
Indeed, the cluster assumptions is a necessary condition for (A1+A4) to hold when the considered space of functions is included in the class of H\"older functions (which is true for all classical reproducing kernel Hilbert space). 
This can be proven with the same derivations as the one above (replacing $\eta$ by $g_\lambda$ and $\eta_0$ by $\delta/2$ with their notations).
In contrast, we do not need regularity of $\eta$, which we show in practice on Figure \ref{svm:fig:fig3}.

\section{NUMERICAL ANALYSIS}

In this section, we provide experiments to illustrate and validate our theoretical findings.
In order to be inline with the current practice of machine learning, instead of considering the hard constraint $\norm{\theta} \leq M$ when minimizing a risk functional, we add a penalty $\lambda \norm{\theta}^2$ to the risk to be minimized.
Going from a constrained to a penalized framework does not change the nature of the statistical analysis, and one might loosely think of $\lambda$ as $1 / M$ \citep[see, for example,][]{Bach2023}.
All experiments are made with the Gaussian kernel.
Precise details of the different settings are provided in Appendix \ref{svm:app:experiments}.

First, we observe that the regime described in this paper kicks in when the error is already pretty small.
On many real-world problems, we do not expect the generalization error as a function of the number of data used for training to exhibit a clear exponential behavior until an unusually big number of samples is used.
This fact is illustrated on Figure~\ref{svm:fig:fig1}, where for hard problems, the exponential behaviors still do not kick in after a thousand of samples.

Second, this paper shows that, in order to get exponential convergence rates for SVM, one needs the minimizer $g_{M,\sigma}$ of the surrogate risk over the selected class of functions to be a perfect classifier, {\em i.e.} its sign equals the sign of $g^*$.
While this is not constraining under the cluster assumption, we inspect divergences from this condition on Figure~\ref{svm:fig:fig2}.
We observe that, even if $g^*$ does not depend on the noise, $g_{M,\sigma}$ does.
We also observe that the regularity of the decision boundary $\brace{x\in \X\midvert \eta(x) = 0}$ should match the regularity defined implicitly by the kernel $k$ and the scale parameter $\sigma$.

\begin{figure*}[t]
	\centering
	\includegraphics{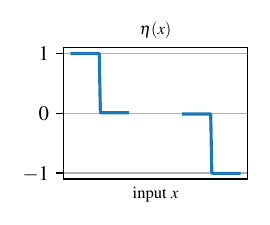}
	\includegraphics{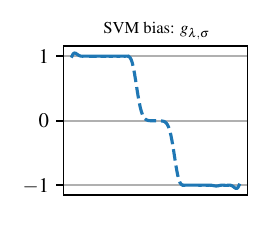}
	\includegraphics{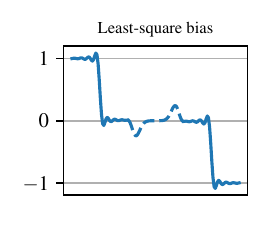}
	\caption{
		Comparison of the regularized risk ({\em i.e.} ${\cal R}_S(g_\theta) + \lambda\norm{\theta}^2$) minimizer for the hinge loss surrogate (middle) and the least-squares surrogate (right), when $\eta$ is not regular (left).
		In this setting, the hinge loss is minimized for $g = \sign(\eta)$, which can be chosen regular, while the least-squares loss is minimized for $g = \eta$, which can not be chosen regular.
		The reconstruction is made with $\sigma$ about 3\% of the domain diameter, and $\lambda$ relatively small.
		We assume no density in the middle of the domain, explaining the absence of definition of $\eta$ and the dashed lines of the right figures.
		The oscillation on the later figures is related to the Gibbs phenomenon \citep{Wilbraham1848}.
		This phenomenon prevents the regularized least-squares solution from being a perfect classifier.
	}
	\label{svm:fig:fig3}
\end{figure*}

Experimental comparisons of different classification approaches have been done by many people, and our goal is not to showcase the superiority of the SVM over least-squares, which might be considered as general wisdom that led to the golden age of SVM in the pre-deep-learning area \citep[see][for example]{Joachims1998}.
In comparison with previous works based on calibration inequalities \citep{Rosasco2004,Steinwart2007b}, our analysis proves the robustness of SVM to noise far away from the decision boundary, in the sense that one does not need $\eta$ to be bounded away from zero.
This is a distinctive aspect of SVM compared to smooth surrogate methods \citep{Nowak2019b}, such as softmax regression, that implicitly estimate conditional probabilities and whose performance depends on the regularity of $\eta$.
We illustrate this fact graphically on Figure~\ref{svm:fig:fig3}.

\section{LIMITATIONS}

\paragraph{Are Surrogate Methods Only a Proxy for Classification?}
From a theoretical perspective, if we are only interested in the optimal mapping $f:\X\to\Y$, learning surrogate quantities can be seen as a waste of resources.
In essence, this waste of resources is similar to the one occurring when we learn the full probability function $(p(y))_{y\in\Y}$ for some probability distribution $p$ on $\Y$, while we only care about its mode.
Yet, in practice, what we call a ``surrogate'' problem might actually be a problem of prime interest when we do not only want to predict $f^*(x)$, but we would also like to know how much we can confidently discard other potential outputs for an input $x$.
Furthermore, assuming that a problem is exactly defined through an ``original'' loss that defines a clear and unique measure of error can be questioned when some practitioners evaluate methods with several metrics of performance \cite[\emph{e.g.}][]{Chowdhery2022}.

\paragraph{Do PAC-Mounds Provide Confidence Levels?}
Since the parameters in Assumptions \ref{svm:ass:weak_margin} and \ref{svm:ass:mass} are hard to estimate in practice, it would be difficult to directly plug our bounds into a practical problem to derive confidence levels on how much error one might expect when deploying a model in production.
Less ambitiously, we see theorems akin to Theorem \ref{svm:thm} as providing theoretical indications that a learning method or a set of hyperparameters is sound.
This is a generic downfall of probably approximately correct (PAC) generalization bounds \citep{Valient2013}, which might explain why practitioners often prefer to derive error indications from test samples \citep[see, \emph{e.g.},][]{Geron2017}.
Along this line, research on conformal prediction provides interesting considerations to obtain useful confidence information from test samples \citep{Vovk2008}.
Finally, all these statistical methods to get confidence intervals assume representative (if not i.i.d.) data, an assumption sometimes hard to meet in practice, which is a problem that has found echoes in the civil society \citep[\emph{e.g.}][]{Benjamin2019}.

\paragraph{Societal Impact}
The theoretical nature of the present work prevents us to discuss its potential negative social impact without questioning the impact of the whole field of machine learning, which is out of scope of the current paper.

\section{CONCLUSIONS}


In this work, we were keen to illustrate a simple mechanism to get exponential convergence rates for support vector machines.
Our proof relates the misclassification to the surrogate risk through a sort of $L^p-L^q$ interpolation inequality.
Thanks to this new strategy,
we were able to deal with a non-smooth loss such as the hinge loss, which is quite popular and whose understanding can not be easily reduced to previous work.
Remarkably, our assumptions are strictly weaker than all the existing assumptions we are aware of used to prove exponential convergence rates.
In particular, we showed that the hard low-noise condition is not crucial in order to derive exponential convergence rates for the SVM.

This provides a crucial step to better understand convergence rates on classification problems.
An extension to generic discrete output problems could be made by considering polyhedral losses, and deriving variants of Lemma~\ref{svm:lem:l1} \citep[see][for calibration inequalities for such losses]{Frongillo2021}.
An important follow-up would be to provide a more global picture of fast polynomial rates for SVM under relaxations of Assumptions \ref{svm:ass:mass} and \ref{svm:ass:source}.

Finally, \cite{Chizat2020} have made a link between two-layer wide neural networks in the interpolation regime (which implies Assumption \ref{svm:ass:margin} with $\eta_0 = 1$) and max-margin classifiers over specific linear classes of functions.
As a consequence, we could directly plug in our analysis to prove exponential convergence rates for those small neural networks in this noiseless setting.
Studying rates, constants and hyperparameter tuning in this setting would be of particular interest if it was to provide practical guidelines to deep learning practitioners in the spirit of \cite{Yang2021}.

\subsubsection*{Acknowledgements}
VC would like to thank Alex Nowak-Vila for sharing insights that led to this work, as well as Francis Bach for useful comments.
SV is partially supported by  the MIUR Excellence Department Project MatMod@TOV awarded to the Department of Mathematics, University of Rome Tor Vergata.
The authors are also deeply grateful for the highly valuable feedback from the anonymous reviewers.

\bibliography{main}

\appendix
\onecolumn
\section{PROOFS}
\label{svm:app:proofs}


In this section we provide the proofs of
Lemma \ref{svm:lem:l1},
Lemma \ref{svm:lem2},
Proposition \ref{svm:prop:mid},
Theorem \ref{svm:thm}
and Proposition \ref{svm:prop:cluster}.

\subsection{Proof of Lemma \ref{svm:lem:l1}}

The first part follows by integration of a pointwise result.
Consider the function $h_p: \R\to\R; q \mapsto p(1-q)_+ + (1-p)(1+q)_+$, where $p\in(0, 1)$ represents $\Pbb(Y=1\vert X)$ and $q$ represents $g(x)$.
The function $h_p$ has a slope equal to $-p$ for $q < -1$, then slope $1-2p$ for $q\in(-1, 1)$, and $1-p$ for $q > 1$. Therefore, when $q_1, q_2 \in (-1, 1)$, we have
\[
	h_p(q_2) - h_p(q_1) = (1-2p) (q_2 - q_1).
\]
Taking $p = \Pbb(Y=1\vert X)$, $q_2 = g_2(X)$ and $q_1 = g_1(X)$, we get $1-2p = -\E[Y\vert X] = -\eta(X)$. By integration, we obtain the claim.
From the previous slope considerations, it also follows that $h_p$ is minimized by $q=\sign(2p-1)$, meaning that one can take $g^*(X) = \sign\eta(X)$.

More exactly, the slopes reasoning shows that: for $x\notin\supp\rho_\X$, $g^*(x)$ and $f^*(x)$ can be arbitrarily chosen; for $x\in\supp\rho_\X$ such that $\eta(x) = 0$, $g^*(x)$ and $f^*(x)$ can be arbitrarily chosen in $[-1, 1]$ and $\brace{-1, 1}$; for $x\in\supp\rho_\X$ with $\eta(x)=1$, $f^*(x)=1$ and $g^*(x)$ can be arbitrarily chosen in $[1, +\infty)$; finally, for $x\in\supp\rho_\X$ and $\eta(x) \in (0, 1)$, $g^*(x) = f^*(x) = 1$.

The second part follows from the fact that projecting on $[-1, 1]$ can only reduce the value of the hinge loss, that $\eta(x)(\pi(g)(x) - g^*(x))$ is always negative, and the reverse H\"older inequality:
\[
	{\cal R}_S(g) - {\cal R}(g^*) \geq {\cal R}_S(\pi(g)) - {\cal R}(g^*) = \norm{\eta (\pi(g) - g^*)}_{1}
	\geq \norm{\pi(g) - g^*}_{q} \norm{\abs{\eta}^{-1}}_{p}^{-1}.
\]
A H\"older inequality also holds for weak Lebesgue spaces \citep[see][Theorem 5.23]{Castillo2016}, whence
\[
	{\cal R}_S(g) - {\cal R}(g^*) \geq \norm{\eta (\pi(g) - g^*)}_{1}
	\geq \norm{\eta (\pi(g) - g^*)}_{1,\infty}
	\geq \frac{1}{2} \norm{ |\eta|^{-1} }_{p,\infty}^{-1} \norm{ \pi(g) - g^* }_{\frac{p}{p+1},\infty}.
\]
This completes the proof.

\subsection{Proof of Lemma \ref{svm:lem2}}
Assume without restrictions that there exists $x \in \X_1$ such that $|g^*(x) - g(x)| \geq 1 $.
For any event $A = A(X)$, by the law of total probability we have
\[
	\Pbb(A) =
	\rho_\X(\X_1) \Pbb\paren{A\midvert X\in \X_1}
	+ \rho_\X(\X_{-1}) \Pbb\paren{A\midvert X\in \X_{-1}}
	\geq \rho_\X(\X_1) \Pbb\paren{A\midvert X\in \X_1}.
\]
Hence, since $g^*(\X_1) = \brace{1}$,
\begin{align*}
	\norm{g - g^*}_{q, \infty}^q
	= \sup_{t > 0} t^q \Pbb(\abs{g(X) - g^*(X)} > t)
	\geq \sup_{t > 0} t^q\Pbb\paren{\abs{g(X) - 1} > t\midvert X\in\X_1}\rho_\X(\X_1).
\end{align*}
Using the triangular inequality, the $G$-Lipschitz continuity of $g$, and the definition of $x$, we have that, for any $x'\in\X$,
\[
	|g(x') - 1| \geq |g(x) - 1| - |g(x') - g(x)| \geq 1 - Gd(x, x').
\]
As a consequence,
\[
	\Pbb\paren{|g(X) - 1| > t\midvert X\in\X_1}
	\geq \Pbb\paren{X\in{\cal B}\paren{x, \frac{1-t}{G}} \mid X\in\X_1}
	= \frac{\rho_\X\paren{\X_1\cap {\cal B}\paren{x, \frac{1-t}{G}}}}{\rho_\X\paren{\X_1}}.
\]
Combined with the previous facts, we get
\[
	\norm{g - g^*}_{q, \infty}^q
	\geq \sup_{t > 0} t^q \rho_\X\paren{\X_1\cap {\cal B}\paren{x, \frac{1-t}{G}}}.
\]
Thanks to Assumption \ref{svm:ass:mass}, there exists $(c, r, d)$ such that \eqref{svm:eq:mass} holds for $\X_1$. Hence, when $G^{-1} < r$, we get the following lower bound:
\[
	\norm{g - g^*}_{q, \infty}^q
	\geq c G^{-d} \sup_{t \in [0, 1]} t^q (1 - t)^d
	= c G^{-d} \frac{q^q d^d}{(d+q)^{d+q}}.
\]
This proves the statement in the lemma.

\subsection{Proof of Proposition \ref{svm:prop:mid}}
Suppose $ {\cal R}(\sign g) > {\cal R}(f^*) $.
Then, observing that $ \sign(\pi(t)) = \sign(t) $ for all $ t \in \R $, and taking $ g^* = f^* $, we know there must be $ x \in \supp \rho_\X $ such that $ \abs{ \pi(g(x)) - g^*(x) } \geq 1 $.
Hence, by Lemma \ref{svm:lem2}, we get $ \norm{ \pi(g) - g^* }_{q,\infty} \geq c_0 G^{-\frac{d}{q}} $, and therefore, by Lemma \ref{svm:lem:l1},
$
	{\cal R}_S(g) - {\cal R}_S(g^*)
	\geq 2^{-1} c_0 G^{-\frac{d}{q}} \norm{\abs{\eta}^{-1}}_{p,\infty}^{-1} .
$
Thus, the proposition is proved.

\subsection{Proof of Theorem \ref{svm:thm}}
From Proposition \ref{svm:prop:mid} and Assumption \ref{svm:ass:source}, we get, with $\tilde{L} = \max\brace{M G_\phi \sigma^{-1}, r^{-1}}$ and $q = \sfrac{p}{p+1}$, and $g_{M,\sigma}$ the minimizer of ${\cal R}_S$ insider $G_{M,\sigma}$,
\begin{align*}
	\E_{{\cal D}_n}[{\cal R}(\sign g_{{\cal D}_n})] - {\cal R}(f^*)
	&\leq \Pbb_{{\cal D}_n} \paren{{\cal R}_S(\pi\circ g_{{\cal D}_n}) - {\cal R}_S(g^*) \geq 2^{-1}\norm{\abs{\eta}^{-1}}_{p,\infty}^{-1} c_0 \tilde{L}^{-\frac{d}{q}}}
	\\&\leq \Pbb_{{\cal D}_n} \paren{{\cal R}_S(\pi\circ g_{{\cal D}_n}) - {\cal R}_S(g_{M,\sigma}) \geq 4^{-1}\norm{\abs{\eta}^{-1}}_{p,\infty}^{-1} c_0 \tilde{L}^{-\frac{d}{q}}}.
\end{align*}
To deal with this last quantity, we proceed by using the fact that
\[
    {\cal R}_{S,{\cal D}_n}(\pi\circ g_{{\cal D}_n}) \leq  {\cal R}_{S,{\cal D}_n}(g_{{\cal D}_n}) \leq  {\cal R}_{S,{\cal D}_n}(g_{M,\sigma}),
\]
where ${\cal R}_{S,{\cal D}_n}$ denotes the empirical surrogate risk,
to deduce that
\[
   {\cal R}_S(\pi\circ g_{{\cal D}_n}) - {\cal R}_S(g_{M,\sigma})
   \leq {\cal R}_S(\pi\circ g_{{\cal D}_n}) - {\cal R}_{S,{\cal D}_n}(\pi\circ g_{{\cal D}_n}) + {\cal R}_{S,{\cal D}_n}(g_{M,\sigma}) + {\cal R}_S(g_{M,\sigma}).
\]
Hence, we get the following union bound
\begin{align*}
	\E_{{\cal D}_n}[{\cal R}(\sign g_{{\cal D}_n})] - {\cal R}(f^*)
    &\leq \Pbb_{{\cal D}_n} \paren{{\cal R}_S(\pi\circ g_{{\cal D}_n}) - {\cal R}_{S, {\cal D}_n}(\pi\circ g_{{\cal D}_n}) \geq 8^{-1}\norm{\abs{\eta}^{-1}}_{p,\infty}^{-1} c_0 \tilde{L}^{-\frac{d}{q}}}
	\\&\qquad+ \Pbb_{{\cal D}_n} \paren{{\cal R}_S(g_{M,\sigma}) - {\cal R}_{S,{\cal D}_n}(g_{M,\sigma}) \geq 8^{-1}\norm{\abs{\eta}^{-1}}_{p,\infty}^{-1} c_0 \tilde{L}^{-\frac{d}{q}}}.
\end{align*}
Regarding the first term, we can reuse the literature on Rademacher complexity for linear models on convex risks \citep{Bartlett2002}, as well as the contraction principle in order to add $\pi$ which is 1-Lipschitz \cite{Maurer2016}, which ensures that
\[
	\E_{{\cal D}_n}\bracket{\sup_{g\in{\cal G}_{M,\sigma}}\abs{{\cal R}_S(\pi\circ g) - {\cal R}_{S, {\cal D}_n}(\pi\circ g)}} \leq M \norm{\phi}_{\infty} n^{-1/2}.
\]
Note that Assumption \ref{svm:ass:mass} implies that $\supp\rho_\X$ is compact, hence, if $\phi$ is Lipschitz-continuous, it is bounded on $\supp\rho_\X$.
This allows us to use McDiarmid inequality to get the same type of bound on the deviation of ${\cal R}_S(g_{{\cal D}_n})$ around its mean.
Let $H({\cal D}_n) = \sup_{g\in{\cal G}_{M,\sigma}} {\cal R}_S(\pi\circ g) - {\cal R}_{{\cal D}_n}(\pi\circ g)$. 
Let us decompose ${\cal D}_n = ((x_1, y_1), \cdots, (x_n, y_n))$. We would like to show that if ${\cal D}_n'$ is equal to ${\cal D}_n$ for each datapoint but for $(x_i, y_i)$ that becomes $(x_i', y_i')$ then $H({\cal D}_n) - H({\cal D}_n')$ is bounded.
We have
\begin{align*}
  H({\cal D}_n) - H({\cal D}_n') 
  &= \sup_{g\in{\cal G}_{M,\sigma}} {\cal R}_S(\pi\circ g) - {\cal R}_{S,{\cal D}_n}(\pi\circ g) - \sup_{g'\in{\cal G}_{M,\sigma}} {\cal R}_S(\pi\circ g') - {\cal R}_{S,{\cal D}_n'}(\pi\circ g')
  \\&\leq \sup_{g\in{\cal G}_{M,\sigma}} {\cal R}_{S, {\cal D}_n}(\pi\circ g) - {\cal R}_{S,{\cal D}_n'}(\pi\circ g)
  \\& = n^{-1}\sup_{g\in{\cal G}_{M, \sigma}} L(\pi\circ g(x_i'), y_i') - L(\pi\circ g(x_i), y_i)
  \leq n^{-1}
\end{align*}
Using McDiarmid's inequality, we get
\[
    \Pbb(H({\cal D}_n) - \E[H({\cal D}_n)] \geq t) \leq \exp(-2nt^2).
\]
In other terms, when adding the control we have on the expectation, we get
\begin{equation}
	\Pbb_{{\cal D}_n}\paren{\sup_{g\in{\cal G}_{M,\sigma}} {\cal R}_S(\pi\circ g) - {\cal R}_{S, {\cal D}_n}(\pi\circ g) > t + M\norm{\phi}_{\infty} n^{-1/2}} \leq \exp\paren{-2nt^2}.
\end{equation}
When $8^{-1}\norm{\abs{\eta}^{-1}}_{p,\infty}^{-1} c_0 \tilde{L}^{-\frac{d}{q}} \geq \norm{\phi}_{\infty} M n^{-1/2}$, this leads to
\begin{align*}
	&\Pbb_{{\cal D}_n} \paren{{\cal R}_S(\pi\circ g_{{\cal D}_n}) - {\cal R}_{S, {\cal D}_n}(g_{{\cal D}_n}) \geq 8^{-1}\norm{\abs{\eta}^{-1}}_{p,\infty}^{-1} c_0 \tilde{L}^{-\frac{d}{q}}}
	\\&\leq \exp\paren{-\frac{n}{8} \paren{8^{-1}\norm{\abs{\eta}^{-1}}_{p,\infty}^{-1} c_0 \tilde{L}^{-\frac{d}{q}} - M\norm{\phi}_{\infty} n^{-1/2}}^2}
	\\&\leq \exp\paren{-\frac{c_0^2 \sigma^{\frac{2d(p+1)}{p}}}{512\norm{\abs{\eta}^{-1}}_{p,\infty}^2(MG_\phi)^{\frac{2d(p+1)}{p}}}\cdot n + \frac{c_0\norm{\phi}_{\infty}}{32\norm{\abs{\eta}^{-1}}_{p,\infty}M^{\frac{d(p+1)}{p}-1}G_\phi^{\frac{d(p+1)}{p}}} \cdot n^{-1/2} - \frac{M^2\norm{\phi}_{\infty}^2}{8}}.
\end{align*}

Regarding the second term, we can use the classical concentration of ${\cal R}_{{\cal D}_n}(g_{M,\sigma})$ around its mean.
For example, using the fact that Assumption \ref{svm:ass:mass} implies that $\rho_\X$ is compact, and using the fact that $L$ and $g_{\sigma, M}$ are Lipschitz, we deduce that $L(g_{\sigma, M}, Y)$ is bounded, hence one can applies Hoeffding's inequality to get the same type of exponential control on this term.

The result follows from those concentration inequalities and the fact that ${\cal R}$ is bounded by one, since any function $h:\N^*\to\R;n\to\min(1, a \exp(-bn))$ given two constants $a, b > 0$ can be bounded by $h':\N^*\to\R;n\to 2\exp(-cn)$ for a constant $c>0$.

\subsection{Proof of Proposition \ref{svm:prop:cluster}}
\label{app:proof_source}

Since the Hinge loss is 1-Lipschitz from the proof of Lemma \ref{svm:lem:l1}, we have that, for any function $g:\X\to\R$,
\[
    {\cal R}_S(g) - {\cal R}_S(g^*) \leq \norm{g - g^*}_{L^1(\rho_\X)} \leq \norm{g - g^*}_{L^2(\rho_\X)}.
\]
To verifies Assumption \ref{svm:ass:source}, it is sufficient to prove that there exists a $\sigma > 0$ such that for any $c, r > 0$, there exists a $M > 0$ and a function $g\in{\cal G}_{M,\sigma}$ such that
\(
    \norm{g - g^*}_1 \leq c M^{-r}.
\)

Under Assumption \ref{svm:ass:cluster}, $g^*$ can be taken as a non-analytic smooth function, {\em e.g.}
\[
    g^*(x) = \frac{\exp(-d(x, \X_{-1})) - \exp(-d(x, \X_1)}{\exp(-d(x, \X_{-1})) + \exp(-d(x, \X_1)}.
\]
In particular, it can be taken as belonging to any Sobolev space.
On the one hand, any kernel that can be written as $k(x, y) = \phi(\norm{x-y}^2)$ for some function $\phi:\R\to\R$ only contains analytic functions in the resulting function space ${\cal G} = \cup_{M,\sigma} {\cal G}_{M,\sigma}$ \citep{Sun2008}, hence some extra work is needed to show that $g^*$ can be approximated well enough from those spaces: more pr\'ecisely, we need to show that the approximation error within ${\cal G}_{M,\sigma}$ decays faster than $M^{-r}$.
On the other hand, as soon as $m \geq (d+1)/2$, the Sobolev space $H^m(\R^d) = W^{m,2}(\R^d)$ is a reproducing kernel Hilbert space. In particular, the case $m=(d+1)/2$ is associated with the exponential kernel $k(x, y) = \exp(-\norm{x-y})$ \citep[see section 7.3.3 in][for example]{Bach2023}, hence one can consider $g^* \in {\cal G}_{M.\sigma}$ for $M$ big enough, which trivially implies Assumption \ref{svm:ass:source}.

\subsection{Proof of Corollary \ref{thm:cor}}
\label{app:thm}

This theorem is a direct application of strong duality.
Because $\theta \to {\cal R}(g_\theta)$ is convex and $\norm{\theta} \leq M$ has a non-empty relative interior, strong duality holds and
\begin{align*}
    \inf_{\theta; \norm{\theta} \leq M} {\cal R}_S(g_\theta)
    &= \sup_{\lambda \geq 0} \inf_{\theta} {\cal R}_S(g_\theta) + \lambda (\norm{\theta}^2 - M^2) 
    \\&= {\cal R}_S(g_{\theta^*}) + \lambda^* (\norm{\theta^*}^2 - M^2)
    \\&= \inf_{\theta} {\cal R}_S(g_\theta) + \lambda^* \norm{\theta}^2 - \lambda^* M^2.
\end{align*}
where $(\theta^*, \lambda^*)$ is the solution of the Lagrangian problem.
In other terms, for any $M > 0$, there exists a $\lambda \geq 0$ such that $\theta_\lambda = \theta_M$, where $\theta_\lambda$ is the minimizer of the regularized problem, and $\theta_M$ is the minimizer of the constrained problem.

Similarly to the proof of Theorem \ref{svm:thm}, one can prove concentration inequality on ${\cal R}_S(g_{{\cal D}_n, \lambda}) - {\cal R}_S(g_{\theta_\lambda})$, which can then be translated into exponential convergence rates on the original problem.

\section{ADDITIONAL CONTEXT ON SVM}
\label{svm:app:hinge_loss}

In this section, we review the geometrical motivation behind support vector machines, as well as their hinge loss characterization.
Suppose that we are given data $(x_i, y_i)_{i\leq n}$. We would like to find a linear separating hyperplane in the features space ${\cal H} \supset \phi(\X)$ between the points $\brace{\phi(x_i) \midvert y_i = 1}$ and $\brace{\phi(x_i)\midvert y_i = -1}$. This can be formulated as
\[
    \begin{array}{ll}
        \text{find} & \theta  \\
        \text{s.t.} & y_i \theta^\top \phi(x_i) > 0 \qquad \forall\,i\leq n.
    \end{array}
\]
For a feasible $\theta$ such that $\norm{\theta} = 1$, one can compute the margin that separates the positive and negative labeled points along the $\theta$-axis. It reads $\min_{y_i=1, y_j=-1} \theta^\top(\phi(x_i) - \phi(x_j))$. One can also compute the minimal displacement that would make a point change of assigned label according to the classification rule induced by $\theta$, it reads $\min y_i \theta^\top \phi(x_i)$.
Hence, among the feasible $\theta$, the most robust to point displacement, is defined through the maximization of the margin between points and the origin along the $\theta$-axis.
\[
    \begin{array}{ll}
        \text{max} & c  \\
        \text{s.t.} & y_i \theta^\top \phi(x_i) > c\norm{\theta} \qquad \forall\,i\leq n.
    \end{array}
\]
Of course, as soon as there is noise, or if the model is not well-specified, this maximization problem is infeasible.
One way to overcome this is to introduce slack variables that act as budget for points that are too close, or on the wrong side of the separating hyperplane,
\[
    \begin{array}{ll}
        \text{max} & \sum_{i\leq n} \xi_i  \\
        \text{s.t.} & \xi_i < 0 \qquad \forall\,i\leq n\\
        & y_i \theta^\top \phi(x_i) > 1 + \xi_i \qquad \forall\,i\leq n.
    \end{array}
\]
This maximization problem can be rewritten as the minimization problem
\[
    \argmin_\theta \sum_{i\leq n} \max\brace{0, 1-y_i\theta^\top\phi(x_i)}.
\]
This is exactly the empirical risk minimization of the hinge with the linear class of function considered in this paper.
Indeed, our work completely forgets about the geometrical point of view of SVM, it uses the classical framework of statistical learning, where a variational objective is provided by a loss $\ell$ and the minimization is done over functions from inputs to outputs.
Interestingly, the maximum margin principle, which was crucial in the introduction of SVM \citep{Vapnik1995}, reappears in Assumption \ref{svm:ass:cluster} under a weaker form: we do not ask for a clear margin between points that have different labels, but for a margin between points where the optimal classifier should be positive and the ones where it should be negative.
The subtle difference resides in the fact that we allow for labeling noise. In particular, we allow for much more labeling noise than previous works that have only shown exponential convergence rates under the hard low-noise condition.

\section{EXPERIMENTAL DETAILS}
\label{svm:app:experiments}

In our experiments, we used the SVM implementation of \cite{Chang2011} through its {\em Scikit-learn} wrapper \citep{Pedregosa2011} in {\em Python}.
We used {\em Numpy} \citep{Harris2020} to reduce our work to high-level array instructions, and {\em Matplotlib} for visualization \citep{Hunter2007}.
Randomness in experiments was controlled with the random seed provided by {\em Numpy}, which we initialized at zero.

\begin{figure*}[ht]
	\centering
	\includegraphics{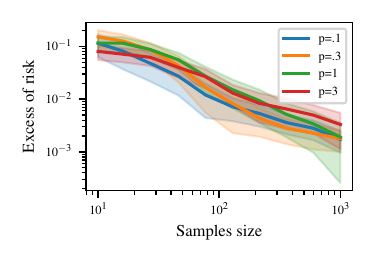}
	\includegraphics{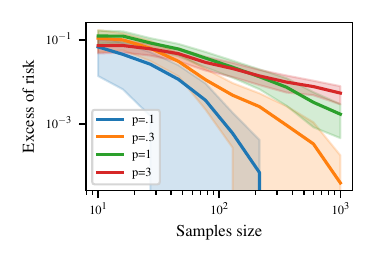}
	\caption{
		(Left) Similar setting as Figure \ref{svm:fig:fig1} but with $X$ uniform on $[-1, 1]$.
		The behavior of the excess of risk is quite different without the separation in $\X$: no exponential convergence rate is kicking in after a thousand of samples.
		(Right) Similar setting as Figure \ref{svm:fig:fig1}, using kernel ridge regression with the least-squares surrogate.
		Exponential convergence rates are observed with a slight delay compared to the hinge loss, and are explained by the hard low-noise condition \ref{svm:ass:margin}.
	}
	\label{svm:fig:fig1_app}
\end{figure*}

Figures \ref{svm:fig:fig1} and \ref{svm:fig:fig1_app} are derived by averaging 100 trials of the following procedure.
We draw uniformly at random $n$ independent samples uniformly distributed on $\X \in \brace{[-1, 1], [-1, -.1]\cup [.1, 1]}$.
We draw randomly one output $y_i$ for each input $x_i$, according to $\eta(x_i)$.
We consider the Gaussian kernel $k(x, x') = \exp(-\norm{x-x'}^2 / 2\sigma^2)$ for $\sigma = .2$, and solve the empirical risk minimization associated to the hinge loss with the penalization $\lambda \norm{\theta}^2$ (rather than the hard constraint $\norm{\theta} < M$) for $\lambda = 10^{-4}$.
The generalization error is measured through the formula $\E[\norm{\eta(x)} \ind{f(x) \neq f^*(x)}]$, with an empirical approximation of this sum with the points $(x_i)_{i\leq n}$ chosen such that $\rho_{\X}([x_i, x_{i+1}]) = 1/n$ and $\rho_{\X}([x_n, +\infty)) = 1/n$, with $n = 10^4$ (which makes sure that the exponential behavior observed is not due to the lack of testing samples).
For each $x$, the height of each dark part corresponds to one standard deviation of the generalization error computed from the 100 trials, and the solid line corresponds to the empirical average.
The fact that the dark parts are not centered around the averages is due to the fact that we have drawn $\log$-plots but centered the interval for linearly-scaled plots.

Figure~\ref{svm:fig:fig2} is obtained by considering $\X = [0, 1]^2$ with uniform input distribution, the Gaussian kernel with $\sigma = .2$, and the penalty parameter $\lambda = 10^{-3}$ (instead of a hard constraint leading to a parameter $M$ as in the main text derivations).
We take $n=10^4 = 100^2$ points uniformly spread out on $\X$ (on the regular lattice $\frac{1}{\sqrt{n}}\cdot \Z^2 \cap \X$) to approximate $g_{\lambda, \sigma}$ with empirical risk minimization on this curated dataset.
We consider $\eta(x) = \pi_{[-1, 1]}(2 x_2 - .5\sin(2\pi x_1) - 1)$, and assign to each $x$ in the dataset a sample $(x, 1)$ weighted by $\Pbb\paren{Y=1\midvert X=x} = (\eta(x) - 1) / 2$, and a sample $(x, -1)$, weighted by $\Pbb\paren{Y=-1\midvert X=x}$.
The ``noiseless'' setting denotes the setting where $\paren{Y\midvert X}$ is deterministic, but with the same decision frontier between the classes $\X_1$ and $\X_{-1}$ characterized by $\brace{(x, .5 + .25\sin(2\pi x))\midvert x\in [0,1]}$.
Once we fit the support vector machine with this dataset, we test it with $n=2.5\cdot 10^5 = 500^2$ data points uniformly spread out on $\X$, and use {\em Matplotlib} to automatically draw level lines.

\begin{figure*}[ht]
	\centering
	\includegraphics{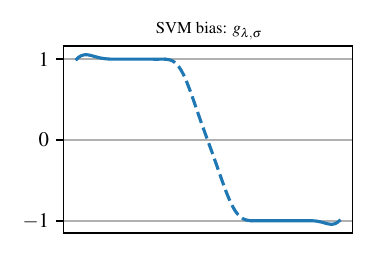}
	\includegraphics{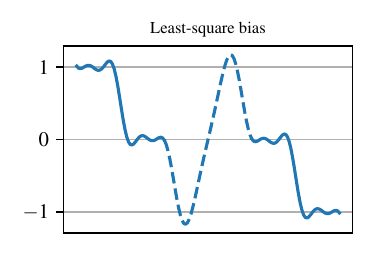}
	\includegraphics{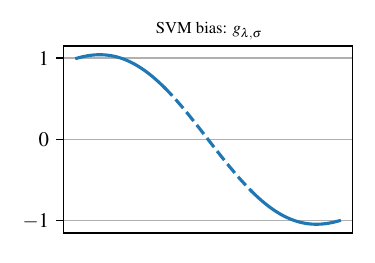}
	\includegraphics{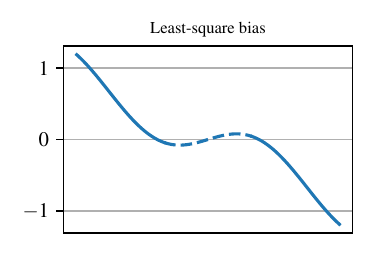}
	\caption{
		Same setting as Figure \ref{svm:fig:fig3}, with $\sigma = .2$ and $\lambda = 10^{-6}$ (top), and with $\sigma = 1$ and $\lambda = 10^{-3}$ (bottom).
	}
	\label{svm:fig:fig3_app}
\end{figure*}

Figures~\ref{svm:fig:fig3} and \ref{svm:fig:fig3_app} correspond to $\X = [0, 3]$ with the input distribution uniform on $[0,1]\cup[2,3]$.
Figure~\ref{svm:fig:fig3} is obtained with $\sigma=.1$ and $\lambda=10^{-6}$.
We derive it by considering $n=100$ points uniformly spread out on the domain of $\eta$, solving the equivalent curated empirical risk minimization, that approximates both
\begin{align}
	g_{\lambda, \sigma} & = \argmin_{g:\X\to\R} \E_{\rho}[(1- Y\scap{\theta}{\phi\paren{\frac{x}{\sigma}}})_+] + \lambda \norm{\theta}^2 ,
	\\g_{(\text{LS})} &= \argmin_{g:\X\to\R} \E_{\rho}[\norm{\scap{\theta}{\phi\paren{\frac{x}{\sigma}}} - Y}^2] + \lambda \norm{\theta}^2.
\end{align}

The robustness of SVM might be understood from its geometrical definition: when trying to find the maximum separating margin, infinitesimal modifications that change the regularity properties of $\eta$ do not really matter.
The picture is different for the least-squares surrogate with kernel methods, where from few point evaluations, the system reconstructs a function by assuming regularity and inferring information on high-order derivatives.
This is similar to the Runge phenomenon with Hermite interpolation.
More precisely, the Gaussian kernel is linked to a space of functions with rapidly decreasing Fourier coefficients \citep[see, for example,][for a more precise link]{Bach2023}.
The function $\eta$ that needs to be approximated on Figure~\ref{svm:fig:fig3} is similar to the Heaviside function, whose Fourier coefficients are of the form $(\frac{1}{i\pi k})_{k\in\N^*}$ and do not decrease fast enough to be all reconstructed.
This leads to some high-frequency oscillations missing in the reconstruction as it appears on Figure~\ref{svm:fig:fig3}.

\fi

\end{document}